\documentclass[conference, a4paper]{IEEEtran}
\IEEEoverridecommandlockouts
\usepackage{amsmath,amssymb,amsfonts}
\usepackage{algorithmic}
\usepackage{graphicx}
\usepackage{textcomp}
\usepackage{xcolor}
\usepackage{booktabs}
\usepackage[backend=biber, style=ieee]{biblatex}
\usepackage{eso-pic}

\begin{document}

\title{MLMC-based Resource Adequacy Assessment with Active Learning Trained Surrogate Models\\
\thanks{This work is part of the project Multi-Energy System Smart Linking (MuESSLi), funded by the Collaborative Research for Energy SYstem Modelling (CRESYM) non-profit association.}
}

\author{\IEEEauthorblockN{Ruiqi Zhang}
\IEEEauthorblockA{\textit{Dept. of Electrical Sustainable Energy} \\
\textit{Delft University of Technology}\\
Delft, The Netherlands \\
r.zhang-3@tudelft.nl}
\and
\IEEEauthorblockN{Simon H. Tindemans}
\IEEEauthorblockA{\textit{Dept. of Electrical Sustainable Energy} \\
\textit{Delft University of Technology}\\
Delft, The Netherlands \\
s.h.tindemans@tudelft.nl}
}

\maketitle

\begin{abstract}
Multilevel Monte Carlo (MLMC) is a flexible and effective variance reduction technique for accelerating reliability assessments of complex power system. Recently, data-driven surrogate models have been proposed as lower-level models in the MLMC framework due to their high correlation and negligible execution time once trained. However, in resource adequacy assessments, pre-labeled datasets are typically unavailable. For large-scale systems, the efficiency gains from surrogate models are often offset by the substantial time required for labeling training data. Therefore, this paper introduces a speed metric that accounts for training time in evaluating MLMC efficiency. Considering the total time budget is limited, a vote-by-committee active learning approach is proposed to reduce the required labeling calls. A case study demonstrates that, within a given computational budget, active learning in combination with MLMC can result in a substantial reduction variance.
\end{abstract}


\section{Introduction}

In response to the increasing variability introduced by higher shares of renewable energy sources (RES) and electrified demand, probabilistic evaluations are widely implemented in power system operation and planning, assessing the impact of uncertainty on system reliability \cite{spilger_uncertainty_2025}. In both administrative and research contexts, two widely used expectation-based metrics are Loss of Load Expectation (LOLE) \cite{stephen_clarifying_2022} and Expected Energy Not Served (EENS) \cite{evans_minimizing_2019}.

Monte Carlo (MC) simulation is a commonly used technique, capable of evaluating the performance of complex systems under multiple sources of uncertainty. However, standard MC simulation suffers from slow convergence, particularly when estimating resource adequacy metrics such as LOLE \cite{stephen_clarifying_2022}, where typical standards are e.g., 3 event-hours in a year or 1 event-day in 10 years. Obtaining high-confidence reliability estimates therefore requires evaluating large number of scenarios. To address this, variance reduction techniques such as importance sampling \cite{liu_assessment_2021} have been developed to alter the sampling strategy to focus on high-impact scenarios samples,  thereby reducing the scenario count, but there is a tight coupling between the simulation model and the available sampling strategies.

Surrogate models, approximating system performance using fitted functions, have been proposed to bypass time-consuming simulations and thereby reduce the computational cost of scenario evaluation \cite{li_surrogate_2019, gao_intelligent_2023}. However, the resulting uncertainty estimates are affected by the prediction errors inherent in surrogate models. Multilevel Monte Carlo (MLMC), as a flexible variance reduction approach, employs a hierarchy of correlated approximating models with different complexity \cite{aslett_multilevel_2017}. Complex models are applied to correct the estimation bias made by reduced models, providing unbiased estimates with uncertainty confidence estimations. When well-trained, surrogate models produce highly correlated outputs with top-level simulations while requiring negligible execution time, making them particularly well suited as lower-level models in the MLMC framework \cite{sharifnia_multilevel_2022, masoumi_deep-learning-enhanced_2024}.

The performance of surrogate models heavily relies on the quantity and distribution of training data. In power system adequacy assessment, however, no pre-labeled scenarios are available, such that each scenario must be evaluated through complex simulations. For large-scale systems, the efficiency gains achieved through surrogate models are often offset by the significant time required for data labeling, which is frequently overlooked in the performance analysis of existing studies.

To mitigate this issue while maintaining accuracy, the number of complex simulation calls should be minimized by maximizing their utility. Active learning offers a potential solution by selectively labeling only the most informative scenarios, so that each optimization call contributes maximally to improving the surrogate model’s accuracy \cite{priesmann_artificial_2024}.

Therefore, this paper considers the total time budget for performing MLMC, including both the time for running Monte Carlo simulations and the time for labeling and training surrogate models. The main contributions of this work are:
\begin{itemize}
    \item The MLMC speed metric from \cite{tindemans_accelerating_2020} is adapted to include the impact of surrogate model training time.
    \item MLMC is combined with a vote-by-committee active learning strategy to improve accuracy for a given training time.
    \item A case study is presented to demonstrate the efficacy of active learning-trained surrogate models in improving the computational efficiency of MLMC-based power system adequacy assessments.
\end{itemize}

\section{MLMC framework with surrogate models}
\subsection{MLMC framework for adequacy assessment}
\begin{figure*}[htbp]
\centerline{\includegraphics[width=0.85\linewidth]{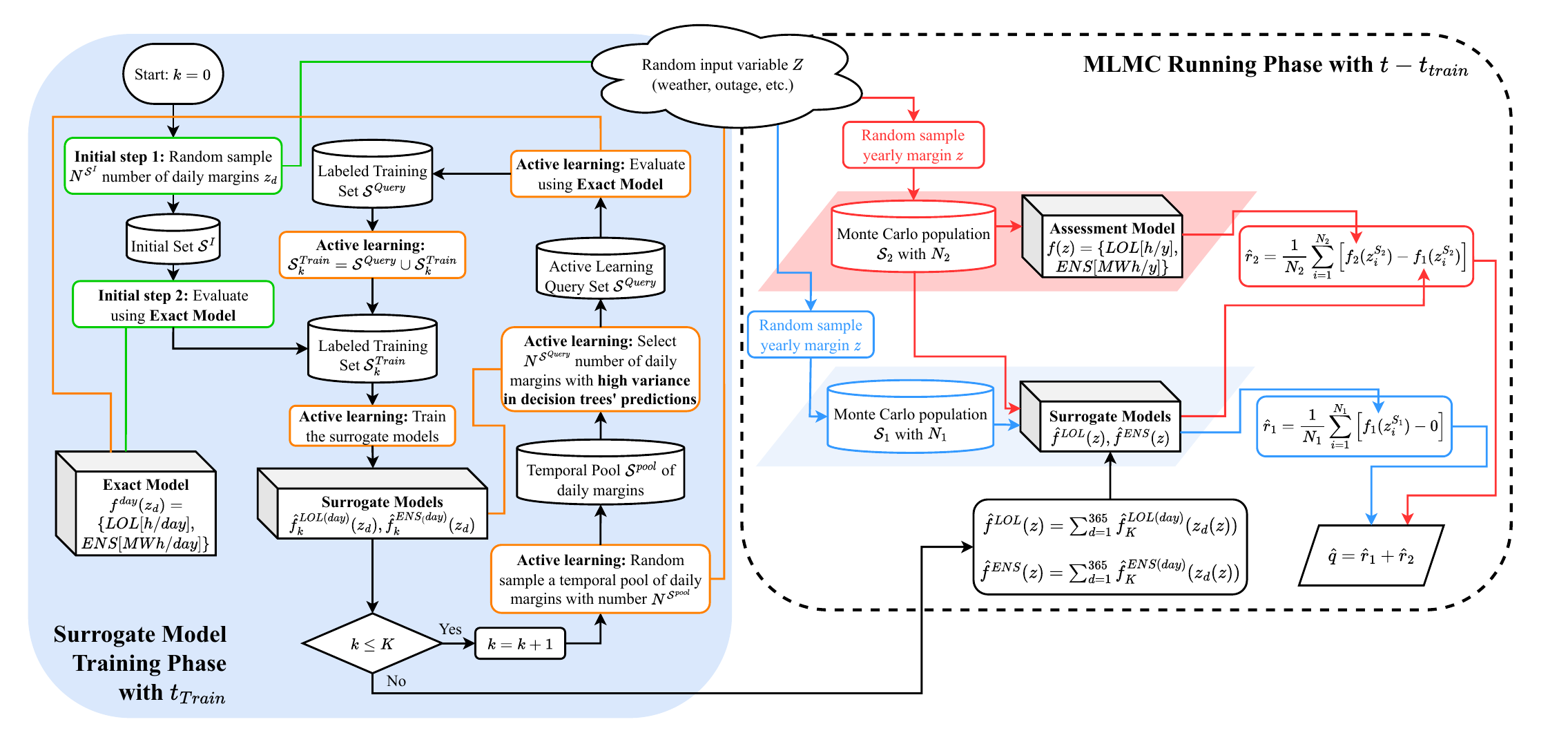}}
\caption{MLMC framework for resource adequacy assessment with AL trained surrogate models. The action flow is marked by black arrows; green and orange boxes with linked models represent initialization and active learning steps, respectively. Red and blue outlines represent the reference and surrogate model evaluations, respectively, in the MLMC process.}
\label{fig:active-learning-surrogate-MLMC}
\end{figure*}
In conventional MC methods, the expectation-based metric  $q=\mathbb{E}\left[f(Z)\right]$ depends on a random input variable $Z$ (weather, outages, etc.) is estimated by directly evaluating a high-fidelity and computationally intensive model $f$ for sampled inputs $z_i$: 
\begin{equation}
    \hat{q}=\frac{1}{N}\sum_{i=1}^{N} f(z_i).
\end{equation}
In contrast, MLMC employs a hierarchy of models $f_1(z),f_2(z),\cdots,f_{L-1}(z)$ with increasing fidelity and complexity to approximate $f_L(z) \equiv f(z)$. These reduced models are much easier to solve, but generally have an approximation error so that $\mathbb{E}\left[f_l(Z)\right] \neq \mathbb{E}\left[f_L(Z)\right]$ \cite{tindemans_accelerating_2020}. Using $f_0 \equiv 0$, the expectation-based metric $q$ can be decomposed into $L$ levels:
\begin{equation}
    \label{eq:MLMC_exact}
    q = \mathbb{E}[f_L(Z)] = \sum_{l=1}^L \mathbb{E}[f_l(Z)-f_{l-1}(Z)].
\end{equation}
By independently generating MC samples at each level, the MLMC estimate for $q$ can be obtained as
\begin{equation}
    \label{eq:MLMC_estimate}
    \hat{q} = \sum_{l=1}^L \frac{1}{N_l} \sum_{i=1}^{N_l}\left[f_l(z^{S_l}_i)-f_{l-1}(z^{S_l}_i)\right].
\end{equation}
Here, $S_l$ and $N_l$ denote the sample set and its size at level $l$, respectively, and $f_{l}(z_i^{S_l})$ represents the output of $f_l$ on a sample $z_i$ over the sampled scenario set $S_l$ at level $l$. 

Interpreting Eq.~\eqref{eq:MLMC_estimate}, the MLMC estimator sums up the output of the lowest level model and the differences between models measured at adjacent levels. The samples at each level are generated independently. In this way, the estimation made by the lowest level model are refined incrementally by the models with higher complexity. A two-layer MLMC methodology framework is shown on the right panel of Fig.~\ref{fig:active-learning-surrogate-MLMC}.

The variance of the MLMC estimator with an optimal allocation of samples for a given simulation time budget $t_{sim}$ is given by \cite{tindemans_accelerating_2020}
\begin{equation}
    \label{eq:MLMC_variance}
    \sigma^2_{\hat{Q}} = \sum_{l=1}^L \frac{\sigma_{Y_l}^2}{N_l} = \frac{1}{t_{sim}}\left(\sum_{l=1}^L \sigma_{Y_l} \sqrt{\tau_l}\right)^2, 
\end{equation}
where $Y_l=f_l(Z^{S_l}) - f_{l-1}(Z^{S_l})$, so that
\begin{align}
    \label{eq:covariance_definition}
    \sigma^2_{Y_l} &= \sigma^2_{f_{l}(Z^{S_l})}+\sigma^2_{f_{l-1}(Z^{S_l})} -2\cdot \text{Cov}(f_{l}\left(Z^{S_l}),f_{l-1}(Z^{S_l})\right).
\end{align}
Here, $\tau_l$ is the time for generating and evaluating the pair of samples at level $l$. The variance decreases with higher correlation between adjacent-level models. Furthermore, although using biased approximations with reduced models, MLMC provides an unbiased estimate of $q$, comparable to that of conventional MC methods, ensuring no loss of accuracy while saving computational resources \cite{sharifnia_multilevel_2022}.

\subsection{Surrogate models for adequacy assessment}
The acceleration achieved by MLMC depends heavily on the correlation between adjacent-level models and their computational cost. Effective lower-level models therefore should be both highly correlated to the upper-level models and computationally more efficient, which often requires case-specific expert knowledge. 

Surrogate models \cite{li_surrogate_2019}, also referred to as meta-models \cite{priesmann_artificial_2024}, have been widely applied to replace heavy simulations in energy system modeling \cite{jiang_surrogate_2022, gao_intelligent_2023, masoumi_deep-learning-enhanced_2024}. To sufficiently capture the actual system behavior across all possible scenarios $Z$, the actual model $f(z)$ must be firstly called over a large number of sampled scenarios to generate a labeled training set. Surrogate models are trained with this training set and once trained, surrogate models generate outputs much more rapidly.

\subsection{Speed measures for MLMC with surrogate models}
The quality of an MC estimate can be summarised by the coefficient of variation of its estimator:
\begin{equation}
    c = \sigma_{\hat{Q}}/q,
\end{equation}
where $\sigma_{\hat{Q}}$ denotes the variance of the estimator and $q$ is the targeted expectation value. The performance of a given MC algorithm, including MLMC, can be characterised by relating $c$ (lower is better) to total computation time $t$. This way, we can identify the computational `speed' (linear in time) of an MC algorithm as \cite{tindemans_accelerating_2020}
\begin{equation}
    \label{eq:speed_MC_def}
    \frac{1}{c^2} = \underbrace{\frac{q^2}{t\cdot \sigma^2_{\hat{Q}}}}_{\text{speed}}\cdot t.
\end{equation}
Combining Eqs.~\eqref{eq:MLMC_variance} and \eqref{eq:speed_MC_def}, and taking $t=t_{sim}$ (all time is spent on simulation), the speed metric for the MLMC framework is
\begin{equation}
    \label{eq:speed_MLMC_surrogate}
    s \equiv \frac{q^2}{t_{sim} \sigma^2_{\hat{Q}}} = \frac{q^2}{\left(\sum_{l=1}^L \sigma_{Y_l} \sqrt{\tau_l}\right)^2}.
\end{equation}
This speed is constant and represents the linear relationship between the estimator performance ($1/c^2$) and total simulation time. Moreover, because $c$ is a relative metric, this speed is independent of the magnitude of $q$, making it suitable for comparing MC estimators across different reliability metrics, such as LOLE and EENS. 

When constructing MLMC with surrogate models, the time $t_{train}$ required to generate, label, and train the surrogate model, should be included in the total time budget $t=t_{train}+t_{sim}$, as the training set is not available a priori. Thus, the speed formulation in Eq.~\eqref{eq:speed_MC_def} is modified as:
\begin{align}
    \label{eq:speed_MLMC_surrogate_full}
    \nonumber
    \underbrace{\frac{1}{c^2}}_{\text{Performance}} &= \underbrace{\frac{q^2}{(t-t_{train})\cdot \sigma^2_{\hat{Q}}}}_{\text{Speed}}\cdot t - \underbrace{\frac{q^2}{(t-t_{train})\cdot \sigma^2_{\hat{Q}}}\cdot t_{train}}_{\text{Training cost of surrogate model}} \\
    &= s \cdot (t-t_{train})
\end{align}
Under a fixed total time budget $t$, the relative performance of two MLMC estimators with different surrogate model training time can be expressed as:
\begin{equation}
    \label{eq:speed_MLMC_comparison_general}
    \frac{1/c^2_a}{1/c^2_b}[t] =\underbrace{\frac{\quad\quad s_a \quad\quad}{s_b}}_{\text{asymptotic speedup}} \cdot \underbrace{\frac{\quad t-t_{train_a} \quad}{t-t_{train_b}}}_{\text{correction for training time}}
\end{equation}
Eq.~\eqref{eq:speed_MLMC_comparison_general} shows that while MLMC with surrogate models may achieve higher asymptotic speed, this advantage is reduced when the available time budget is limited. The effective speedup becomes a function of both the asymptotic speedup and the fraction of time consumed by training.
\section{Active learning strategy for training surrogate models}
The performance of data-driven surrogate models heavily depends on the labeled data used for training. This includes both the quantity of data samples and their informativeness, which contribute to the model’s generalization ability on unseen real-world scenarios \cite{li_active_2024}. In the application of adequacy assessments, historical labeled dataset is often limited or entirely unavailable for training. Each data point must therefore be labeled through computationally expensive operation models. Suffering from the high computation burden, the number of samples that can be evaluated is thus constrained, making the effective selection of training samples crucial.

An active learning (AL) approach is therefore proposed to selectively label on the samples with high informativeness, maximizing the contribution of each data point in improving the generalization ability and therefore saving computation time for expensive labeling simulation \cite{settles_active_2009}.

Instead of directly labeling a large set of randomly chosen scenarios, a small initial subset is labeled to train an initial surrogate model. In subsequent AL iterations, a pool of unlabeled scenarios is generated and their outputs are predicted by the trained model. Based on a predefined query strategy, the most uncertain samples are identified and selected for labeling via the exact model. Two commonly applied AL strategies include:
\begin{itemize}
    \item \textbf{Vote-by-committee (disagreement-based) \cite{bamdad_building_2020}}: An ensemble of surrogate models is trained, e.g. through bootstrapping of the training data and randomizing initial conditions.  Each model is used to predict outcomes for the unlabeled scenarios and the degree of disagreement (e.g. measured by the variance) is used to select points with output uncertainty. 
    \item \textbf{Uncertainty sampling (confidence-based) \cite{mahmood_active_2024}}: For surrogate model types that explicitly provide confidence estimates, scenarios with low confidence are selected for evaluation.
\end{itemize}
After new labeled data points are obtained through the querying process, the surrogate models are retrained with the updated training set, and the active learning loop continues until reaching the predefined iteration time $K$. It is anticipated that targeting scenarios with high uncertainty will rapidly improve the quality of prediction and increase the covariance in Eq.~\eqref{eq:covariance_definition}.

\section{Case study: Single-node generation adequacy assessment}
\subsection{System description}
To investigate the impact of surrogate model training time on MLMC performance when surrogate models are used as lower-level models, a case study of resource adequacy assessment is implemented on a single-node power system (also used in \cite{sharifnia_multilevel_2022}). The system includes aggregated loads, wind turbines, 12 thermal generators, and 27 storage units operated under an EENS-minimizing dispatch policy \cite{evans_minimizing_2019}. The exact model $f(z)$, based on simulation, evaluates the system’s loss-of-load (LOL) and energy not served (ENS) using a one-year generation margin time-series $z$ with 1 hour resolution:
\begin{equation}
    z[t] = \sum P_g[g,t] + P_w[t] - P_d[t],\ t\in \{1,\cdots,8760\}.
\end{equation}
Here $P_g$ denotes the time-series of available capacities of thermal generators, sampled with 90\% availability. The yearly wind generation $P_w$ and yearly demand $P_d$ are randomly drawn from 30 years of historical wind profiles and 10 years of demand profiles, respectively.

\subsection{Surrogate model trained via active learning}
Two random forest regressors, $\hat{f}^{LOL(day)}$ and $\hat{f}^{ENS(day)}$ are built to estimate daily LOL [h/day] and ENS [MWh/day] based on 24-hour\footnote{We note that training a model on shorter traces will necessarily introduce a bias in the surrogate model, but this is corrected within the MLMC framework.} margin time-series $z_{d}$. Random forests are widely applied ensemble models, naturally consisting of a committee of decision tree models for active learning, providing robust prediction while inexpensive to train \cite{pham_predicting_2020}. The final estimates for the one-year LOL and ENS are computed by aggregating the prediction across 365 days:
\begin{equation}
    \hat{f}(z) = \sum_{d=1}^{365}\hat{f}^{(day)}(z_{d}(z)).
\end{equation}

The vote-by-committee approach was implemented as the active learning strategy for selecting informative samples, as demonstrated on the left panel of Fig.~\ref{fig:active-learning-surrogate-MLMC}. At the initial step, 730 daily margin time series were randomly generated and labeled using the exact model. Entering the active learning phase, given that storage units follows the EENS-minimizing dispatch policy, the ENS estimator was used to predict ENS values for a temporary pool of 3650 new daily margin traces. For each candidate trace, the standard deviation of predictions across individual decision trees in the ENS random forest was computed. The 91 days with the highest standard deviation (most disagreement) were then selected. Their true LOL and ENS values were obtained through simulation and added to the training set for the next iteration. For benchmarking, surrogate models were also trained using training sets constructed by random selecting daily margin traces.

This study uses a basic setting of the random forest regressor, with 100 decision trees and vote-by-committee active learning. While these settings are sufficient to demonstrate the concept, surrogate model performance could be enhanced through hyperparameter tuning and exploration of alternative active learning methods. Moreover, the efficiency of the surrogate-assisted MLMC framework could be further improved through the use of importance sampling strategies.

All simulations were repeated for 10 times and the averaged values are presented in Section \ref{section:Results}. Code was written in Python and executed on a 14-inch MacBook Pro with an Apple M1 Pro processor (16 GB RAM). All code and data are available for download.\footnote{Zenodo, May 27, 2025. doi: 10.5281/zenodo.16600719.}
\section{Results}
\label{section:Results}
To evaluate the performance of both the daily LOL and ENS estimators and the final surrogate models in the MLMC framework, two test sets are generated: (1) \textbf{a daily test set} with 100,000 randomly sampled days, and (2) \textbf{a yearly test set} with 1,000 randomly generated 8760-hour margin scenarios. 

\begin{figure}[bhp]
    \centering
    \includegraphics[width=\columnwidth]{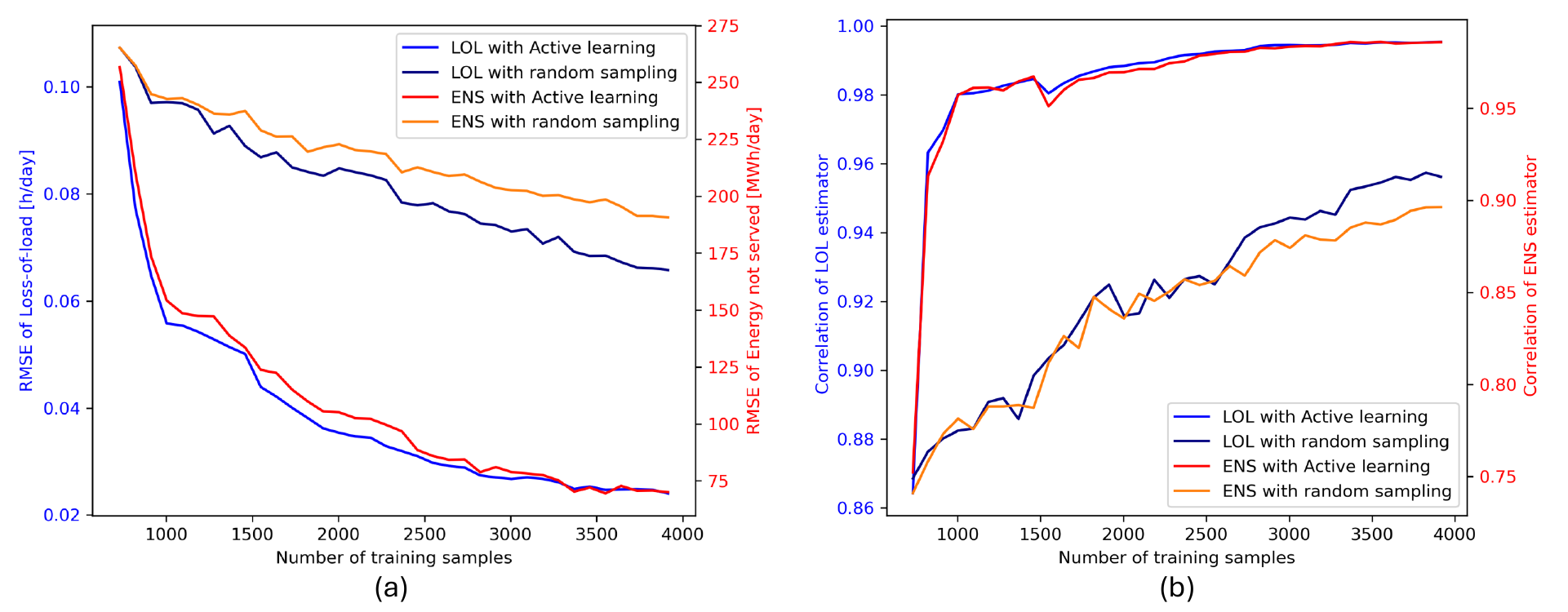}
    \caption{Performance of surrogate LOL and ENS estimators as function of the \emph{number of training samples}. (a) RMSE of 24-hour margin traces; (b) correlation between $\hat{f}(x)$ and $f(x)$.}
    \label{fig:surrogate_performance_before_MLMC}
\end{figure}

\begin{figure}[bhp]
    \centering
    \includegraphics[width=\columnwidth]{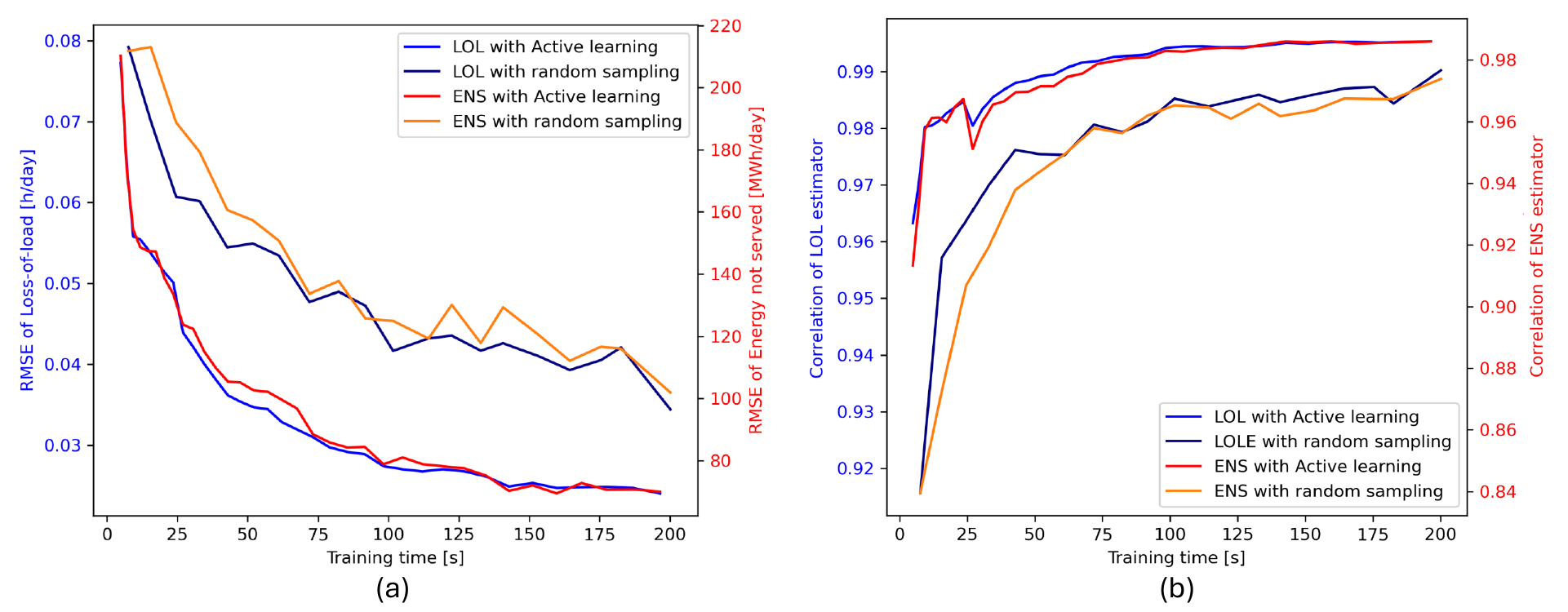}
    \caption{Performance of surrogate LOL and ENS estimators as function of \emph{training time}. (a) RMSE of 24-hour margin traces; (b) correlation between $\hat{f}(x)$ and $f(x)$.}
    \label{fig:surrogate_performance_before_MLMC_time}
\end{figure}

Fig.~\ref{fig:surrogate_performance_before_MLMC} shows the surrogate model accuracy as function of the number of training samples. For a given training set size, the surrogate models trained using active learning consistently show lower RMSE and higher correlation with the reference model than those trained via random sampling. Since active learning introduces additional computational overhead from iterative evaluation and retraining, Fig.~\ref{fig:surrogate_performance_before_MLMC_time} further compares model performance as a function of training time (including data generation, labeling, and model training). Consistently, surrogate models trained with active learning outperform those trained with random sampling in both accuracy and efficiency.

\begin{table*}[htbp]
    \caption{MLMC framework comparison between surrogate models trained with active learning and random sampling}
    \centering
    \begin{tabular}{p{2.3cm}|p{1.2cm}p{0.9cm}p{1.1cm}|p{1.6cm}p{0.9cm}p{1.3cm}|p{1.9cm}p{0.9cm}p{1.3cm}}
    \midrule \midrule
       Estimator & Train size [days] & Train time [s] & Simulation time [s] & LOLE [h/y] & LOLE speed & Break-even $1/c^2$ & EENS [MWh/y] & EENS speed & Break-even $1/c^2$\\
    \midrule
        Exact model & N/A & 0 & 2004 & $1.753 \pm 0.146$ & 0.118 & N/A & $2417.2\pm 209.0$ & 0.067 & N/A\\
    \midrule
        AL (5 rounds) & $1185$  & 13.53 & 2004 & $1.749\pm 0.058$ & 0.829 & 1.86 & $2454.6\pm 136.3$ & 0.296 & 1.18\\
        AL (10 rounds) & $1640$ & 30.65 & 2004 & $1.755\pm 0.034$ & 1.462 & 32.77 & $2442.7\pm 87.5$ & 0.526 & 11.59\\
        AL (20 rounds) & $2550$ & 72.67 & 1999 & $1.739\pm 0.028$ & 2.009 & 225.63 & $2369.0\pm 60.6$ & 0.891 & 53.95\\
    \midrule
        Random surrogate & 3285 & 14.58 & 1996 & $1.724\pm 0.050$ & 0.694 & 2.07 & $2393.2\pm 140.8$ & 0.192 & 1.51 \\   
         & 7300 & 33.59 & 2004 & $1.743\pm 0.039$ & 1.158 & 62.02 & $2396.6\pm 87.4$ & 0.473 & 6.14 \\
         & 14600 & 72.24 & 2000 & $1.742\pm 0.037$ & 1.232 & 745.14 & $2411.6\pm 89.6$ & 0.468 & Invalid \\
    \midrule \midrule
    \end{tabular}
    \label{tab:MLMC_performance results}
\end{table*}

With surrogate models as the lower-level models, the MLMC estimators are further investigated in Tab.~\ref{tab:MLMC_performance results}, comparing models trained with different rounds of active learning and different sizes of randomly sampled training data. In all cases, surrogate-based MLMC significantly reduces computational cost relative to conventional Monte Carlo simulation with the exact model. While simulation speed improves with larger training sets, this gain comes with increased training overhead. With comparable training time, surrogate models trained via active learning consistently achieve higher efficiency than those trained with random sampling.

To assess the cost-effectiveness of model refinement, the break-even performance $1/c^2$, in terms of the target coefficient of variation is calculated. This value represents the point at which the given model becomes more efficient than the one-level-simpler alternative (or the exact model, for the small training sets) in Tab.~\ref{tab:MLMC_performance results}. For example, the LOL surrogate trained over 20 rounds of AL achieves a break-even performance of 225.63 ($c \approx 6.7\%$) relative to its 10-round counterpart. Thus, if the desired estimation variance corresponds to $c < 6.7\%$, it is computationally efficient to apply 20 rounds of active learning, instead of 10 rounds.

In contrast, for surrogate models trained using random sampling, increasing the training size from 7,300 to 14,600 days yields only marginal improvement in LOL accuracy and virtually no improvement in ENS accuracy, despite doubling the training cost. Notably, the LOL estimator trained with 14,600 samples reaches a break-even point of 745.14 ($c \approx 3.7\%$) compared to the 7300-sample case, indicating that additional labeling offers little benefit at practical variance thresholds.

\section{Conclusions and future work}

This paper investigates the impact of surrogate model training time on the overall MLMC performance, explicitly considering the total time budget. A modified speed metric is proposed, showing that the asymptotic speedup achieved through surrogate models is offset by a factor dependent on both the training time and the total time budget. To better utilize the training effort, a vote-by-committee active learning strategy using random forest regression is employed to reduce the number of simulations for labeling. One case study of resource adequacy assessment shows that active learning leads to improved surrogate accuracy and greater MLMC efficiency compared to random sampling. Future work will focus on extending studies to large-scale applications, and further optimizing the labeling process by leveraging information from the high-fidelity model during MLMC execution.

\renewcommand*{\bibfont}{\footnotesize}
\printbibliography

\end{document}